\let\origvec\vec
\documentclass{llncs}

\let\vec\origvec
\usepackage{makeidx}
\usepackage{graphicx}
\usepackage{graphicx, array}
\usepackage{mathtools}

\usepackage{url}
\makeindex
\begin{document}
\title{Linking Tweets with \\ Monolingual and Cross-Lingual News \\ using Transformed Word Embeddings}
\author{Aditya Mogadala\inst{1}, Dominik Jung\inst{2} \and Achim Rettinger\inst{1}}
\institute{Institute AIFB, Karlsruhe Institute of Technology, Germany, \email{aditya.mogadala@kit.edu}, \email{rettinger@kit.edu}
\and
Institute IISM, Karlsruhe Institute of Technology, Germany, \email{dominik.jung2@kit.edu}}
\maketitle

\begin{abstract}
  Social media platforms have grown into an important medium to spread information about an event published by the traditional media, such as news articles. Grouping such diverse 
  sources of information that discuss the same topic in varied perspectives provide new insights. But the gap in word usage between informal social media 
  content such as tweets and diligently written content (e.g. news articles) make such assembling difficult. In this paper, we propose a transformation framework to bridge the word 
  usage gap between tweets and online news articles across languages by leveraging their word embeddings. Using our framework, word embeddings 
  extracted from tweets and news articles are aligned closer to each other across languages, thus facilitating the identification of similarity between news articles and tweets. 
  Experimental results show a notable improvement over baselines for monolingual tweets and news articles comparison, while new findings are reported for cross-lingual comparison.
\end{abstract}
\section{Introduction}
On the web, growth of social media platforms has offered numerous opportunities with several challenges to solve. Twitter~\footnote{\texttt{https://twitter.com/}} is one such social 
media platform that allows its users to share 140 characters of text messages (popularly known as \textbf{tweets}) in multiple languages with their friends or 
followers. Tweets may contain personal information or a confined description about an event motivated by the traditional media such as online news articles. Studies~\cite{kwak:2010} 
have shown that 85\% of the tweets are news affiliated. Though only some tweets acknowledge news articles by explicitly linking them, most of them do not. This implicit linking of 
tweets with the news topics provide novel insights. For example, most of the traditional media companies that publish online news write only facts about an event. However, 
identifying relevant tweets for the corresponding news will append people opinion. Furthermore, attaching tweets with the news articles will allow to understand the 
multi-dimensional view about controversial topics, thus empowering the editor of an article to modify upcoming or following articles based on veracity. 

Howbeit, due to the differences in word usage across informal tweets and the attentively drafted writings like news articles make this linking 
challenging. Nevertheless, different approaches are pursued to solve the problem. Initially, monolingual comparison of tweets with news articles is achieved by 
comprehending commonality between the topics using unsupervised topic models~\cite{zhao:2011}. Although a scalable approach, it fails to capture importance of words and their 
differences across corpora. A graph-based latent variable model~\cite{guo:2013} was further introduced for finding short text correlations using microblog hashtags and news articles 
named entities. Even though it addresses earlier drawbacks by giving importance to keywords such as named entities in news articles. It still ignores other large chunk of 
vocabulary. Krestel et al.~\cite{krestel:2015} followed a different path by posing the comparison of tweets with news as relevance assessment problem and designed 
supervised binary classifier with many hand-crafted features. Yet supervised, the hand-crafted features limit its scalability. Further, aforementioned approaches 
ignore the multilingual aspect of the published news. Nowadays most of the online news about any event is multilingual. Identification of a news article belonging to single language is not enough to cover the collective views about an event. 

In this paper, we overcome the limitations of earlier approaches and propose a new scalable framework to support tweets with monolingual and cross-lingual news article comparison. 
Our framework leverages monolingual~\cite{pennington:2014} and bilingual~\cite{gouws:2014} word embeddings acquired from tweets and news articles as basic units for 
bridging the word usage gap across these collections. Furthermore, non-linear transformation of tweet word embeddings is performed to make it closer to the news article word 
embeddings using manifold alignment with Procrustes analysis~\cite{wang:2008}. Work closely related to our approach is by Tan et al.~\cite{tan:2015} who perform lexical comparison of words observed in tweets and Wikipedia belonging to same language with only linear 
transformation, while we perform non-linear transformation and also across languages. Three main contributions are summarized as follows:
\begin{enumerate}
\item Proposed an approach to classify tweets as to how relevant they are for a given news article in more than one language.
\item New evaluation corpora is created for monolingual and cross-lingual tweets to news article comparison.
\item Lexical and task specific evaluation results are presented on two different datasets.
\end{enumerate}
\section{Related Work}
Most of our research is closely related to the work that identifies relevance of tweets with online news or perform event detection. We divide each of the related works into separate 
categories.
\subsection{Event Detection in Tweets}
Analyzing information flow about the events as they emerge is an important aspect of event detection in tweets. Several works used this information in various ways. Some 
approaches~\cite{ritter:2012} collocated emerging events and classified them into different categories, while some~\cite{thelwall:2011} found sentiment from the detected events. Others 
detected events as trends to track public health~\cite{paul:2011}, political abuse~\cite{ratkiewicz:2011} and crisis communication~\cite{crooks:2013}.
\subsection{News and Relevant Tweets}
Several approaches have been explored to identify relevant noisy tweets with the lengthy news articles. Initially, a semantic enrichment framework~\cite{abel:2011} was built to link 
news articles and tweets by identifying possible correlations to provide personalized news recommendations. Jin et al.~\cite{jin:2011} viewed the problem from different perspective and 
introduced a dual latent Dirichlet allocation model to jointly learn two sets of topics. Later, a more sophisticated unsupervised topic modeling~\cite{zhao:2011} approach was proposed for 
finding overlap of topic distribution between tweets and news articles obtained from New York Times\footnote{\texttt{http://www.nytimes.com/}}. 
\subsection{Distributed Representations}
Distributed word representations~\cite{pennington:2014} has shown significant improvements in many NLP tasks~\cite{coll:2011}. Different variations of them such as 
bilingual~\cite{gouws:2014} and polylingual~\cite{alrfou:2013} are also obtained by projecting multiple or pair of languages into the shared semantic space. Also, word 
representations were extended to meet requirements of the short or noisy text~\cite{ramon:2015,kim:2015}.

\section{Monolingual Word Usage Characteristics}
\label{sec:mwc}
To understand the characteristics of word usage, initially news articles in German and English are collected between January, 2015 and December, 2015. To have a good overlap of 
topics, keywords\footnote{\texttt{https://github.com/aneesha/RAKE}} are extracted from news articles to be used as queries for collecting tweets belonging to the same period with 
Twitter search API\footnote{\texttt{https://dev.twitter.com/rest/public/search}}. Acquired tweets are then polished by removing URLs, user mentions, ``\#'' symbol of the hashtags, 
and all re-tweets. Additionally, Glove\footnote{\texttt{https://github.com/stanfordnlp/GloVe}} is used to obtain word embeddings with 400 dimensions for 
both collections. Size of the final document sets and the vocabulary extracted from Glove is listed in the Table~\ref{cs}.
\begin{table}[!ht]
 \small
\begin{center}
\begin{tabular}{llll}
\hline
Collection	  &     Language     &   Documents  &  Vocabulary            \\ 
\hline
News      &     English      &   1027987    &  348419    \\
News      &     German       &   198784     &  241014    \\
Tweets    &     English      &   110731     &  47280     \\
Tweets    &     German       &   56957      &  31887     \\
\hline
\end{tabular}
\end{center}
\caption{\label{cs}Collection Sizes}
\end{table}

Word embeddings for each collection are now used to effectively comprehend the word usage characteristics. Initially, top 10 common and frequent words 
observed in both collections are visualized with t-SNE~\cite{van:2008}. We observed that the same words learned separately from tweets and news collection are highly separated.
Furthermore to apprehend the difference in slangs, abbreviations etc., in both collections, we use frequent 5000 common vocabulary terms (both English and German) to perceive 
differences among their nearest neighbors. Based on rank biased overlap (RBO) measure~\cite{webber:2010,tan:2015} which provides a comparison between incomplete and indefinite 
rankings, we observe a minimal average RBO measure of $0.2856$ and $0.2589$ for English and German respectively with parameters $\varphi=0.9$ and $k=100$. Thus exhibiting the 
difference in word usage among both collections. This motivates us to transform word embeddings learned with tweets closer to word embeddings learned using news articles or vice versa. 
\section{Transformed Word Embeddings (TWE)}
Difference in the embeddings learned from two different collections such as tweets and news require bridging with embedding transformation. In this section, we formulate the problem and 
present our approach for monolingual and cross-lingual transformation.
\subsection{Problem Formulation}
Let, $T^l_{w_{n}}=\{t^l_{w_1},t^l_{w_2}...t^l_{w_i}...t^l_{w_n}\}$ and $T^l_{e_{n}}=\{t^l_{e_1},t^l_{e_2}...t^l_{e_i}...t^l_{e_n}\}$ represent set of words and their corresponding embeddings 
extracted from tweet collection respectively. Where $l$ is the language of tweets, $n$ is the size of vocabulary and each embedding is of 
dimension $t^l_{e_i}\in R^{1Xd}$. Similarly, $N^l_{w_{m}}=\{n_{w_1}^l,n_{w_2}^l...n_{w_i}^l...n_{w_m}\}$ and $N_{e_{m}}^l=\{n_{e_1}^l,n_{e_2}^l...n_{e_i}^l...n_{e_m}^l\}$ represent set of 
words and their embeddings of news corpora respectively. Where $l$ is the language of news corpora, $m$ is the size of vocabulary and each embedding is of 
dimension $n_{e_i}\in R^{1Xd}$.

Formally, now our research question is to identify common words $\{T^l_{w_{c}},N^l_{w_{c}}\}=\{t^l_{w_i},n_{w_i}^l\}_{i=1}^{c}$ and transform word embeddings in the tweet 
collection ($T^l_{e_{c}}$) closer to the embeddings of news collections ($N^l_{e_{c}}$) or vice versa. This transformation is based on the assumption that there prevails a transformation 
relationship between the vectors for the frequent words of each collection. Some approaches~\cite{tan:2015} have earlier performed this simple transformation only if the language of tweets and 
formal language corpora (e.g. news, Wikipedia) belong to same language. But, it is non-trivial if the language of tweets and formal language corpora differs. 

In the following sections, we present the transformation of tweet embeddings closer to the monolingual or cross-lingual news embeddings. 
\subsection{Monolingual-TWE}
\label{ssec:monot}
Earlier approaches~\cite{tan:2015} assume only linear relationship between embeddings from different collections to perform transformation. Sometimes relationship needs to handle 
disturbances such as scaling and rotation. To cater such issues, we leverage manifold alignment using Procrustes analysis~\cite{wang:2008} to transform word embeddings of tweets 
closer to word embeddings of news articles with a three step procedure.

\begin{itemize}
 \item Learning low-dimensional embeddings is cue for transformation. We already have low-dimensional embeddings $\{T^l_{e_{c}},N^l_{e_{c}}\}$ of words observed in both tweet and 
 news collection.
 \item To find the optimal values of transformation, Procrustes superimposition is done by translating, rotating and scaling the objects (i.e. rows of $T^l_{e_c}$ is 
 transformed to make it similar to the rows of $N^l_{e_c}$). Transformation is achieved by
 \begin{itemize}
  \item \textbf{Translation:} Taking mean of all the members of set to make centroids 
  ($\sum_{i=1}^c\frac{T^l_{e_i}}{c},\sum_{i=1}^c\frac{N^l_{e_i}}{c}$) lie at origin.
  \item \textbf{Scaling and Rotation:} The rotation and scaling that maximizes the alignment is given by orthogonal matrix ($Q$) and scaling factor ($j$). They are obtained by 
  minimizing orthogonal Procrustes problem~\cite{schonemann:1966} and is provided by Equation~\ref{eqn:procru}.
  \begin{equation}
  \label{eqn:procru}
    \arg\min_{j,Q}||N^l_{e_c}-T^{l}_{e^{*}_c}||_F     
  \end{equation}
  where $T^{l}_{e^{*}_c}$ a matrix of transformed $T^{l}_{e_c}$ values given by $jT^{l}_{e_c}Q$ and $||.||_F$ is the Frobenius norm constrained over $Q^TQ=I$.
 \end{itemize}
 \item If $T^{l}_{w^{*}_c}$ represents the words of $T^{l}_{e^{*}_c}$ low-dimensional embeddings, then the final sets $\{T^{l}_{w^{*}_c},N^l_{w_{c}}\}$ contains closer 
 correspondence.
\end{itemize}
To understand the effectiveness of this transformation, we perform similar experiments as of \S~\ref{sec:mwc} in \S~\ref{ssec:monoc}.
\subsection{Cross-Lingual-TWE}
Comparison of vocabulary obtained from tweets in one language ($l_1$) with the vocabulary of news articles in another language ($l_2$) is not straightforward. To subdue 
this concern, we propose a two step approach.
\begin{itemize}
 \item In the first step, news articles from two different languages are acquired to learn bilingual word distributed representations(i.e. bilingual embeddings). Aim of bilingual 
 embeddings is to capture linguistic regularities across languages into a common semantic space such that English and German words (e.g. ``wonderful'' and ``wunderbar'') are neighbors in 
 the t-SNE visualization, thus bridging the language gap. 
 \item In the second step, cross-lingual transformation is achieved between word embeddings obtained from tweets in $l_1$ and word embeddings of news articles in $l_2$. As bilingual word 
 embeddings of news articles in $l_1$ also share linguistic regularities from $l_2$, mapping word embeddings of tweets closer to the bilingual word embeddings of news articles of $l_1$ will also 
 help to incorporate linguistic regularities of $l_2$. Consequently, transformation is attained in the similar way as \S~\ref{ssec:monot} between word embeddings of tweets and 
 bilingual word embeddings of news articles belonging to same language.
\end{itemize}
\subsubsection{Step-1}
To learn bilingual embeddings, we leverage the approach of Gouws et al.~\cite{gouws:2014} as it is fast and scalable to jointly optimize the monolingual objective $M(\cdot)$ with the cross-lingual 
objective $\varphi(\cdot)$ (i.e. cross-lingual regularization term) to find the overall loss $L(\cdot)$. Documents in the news collection of languages $l_1$ and $l_2$ are used to learn 
monolingual models along with cross-lingual regularization term learned with parallel corpora (e.g. Europarl-v7). Overall loss function $L(\cdot)$ is given by Equation~\ref{eqn:loss2}.
\begin{equation}
\label{eqn:loss2}
 L(\cdot) =  \min\limits_{\theta^{l_1},\theta^{l_2}} \sum_{l \epsilon\{l_1,l_2\}} \sum_{C^{l}} M^{l}(w_t,h;\theta^{l}) + \frac{\lambda\varphi(\theta^{l_1},\theta^{l_2})}{2}
\end{equation}
$\varphi(.)$ eliminates the need for word-alignment and makes an assumption that each word observed in the document of language $l_1$ can potentially find its alignment in the 
document of language $l_2$. Thus, the Equation~\ref{eqn:loss2} is now modified into Equation~\ref{eqn:bipd}.
\begin{equation}
\begin{split}
\label{eqn:bipd}
& L(\cdot) =  \min\limits_{\theta^{l_1},\theta^{l_2}} \sum_{l \epsilon\{l_1,l_2\}} \sum_{C^{l}} M^{l}(w_t,h;\theta^{l}) \\
& + \frac{\lambda||\frac{1}{m}\sum\limits_{w_i \epsilon l_1}^{m} V_i^{l_1}-\frac{1}{n}\sum\limits_{w_i \epsilon l_2}^{n} V_i^{l_2}||^2}{2}
\end{split}
\end{equation}
Where $V^{l_1}$ and $V^{l_2}$ are monolingual word vectors of the words in documents of languages $l_1$ and $l_2$ respectively and $C^l$ is monolingual corpus (e.g. News). $w_t$ is the predicted 
word in the context $h$ of a monolingual model.
\subsubsection{Step-2}
We follow a similar procedure as of \S~\ref{ssec:monot} but with a different set of embeddings. 
\begin{itemize}
 \item Low-dimensional embeddings that are used initially are $\{T^{l_1}_{e_{c}},N^{l_1}_{e_{c}}\}$ of words observed in both tweet and news collection belonging to the same 
 language. Here, $N^{l_1}_{e_{c}}$ represents \textbf{bilingual embeddings}.
\end{itemize}
Transformation is now achieved by translating, rotating and scaling the objects (i.e. rows of $T^{l_1}_{e_c}$ is transformed to make it similar to the rows of $N^{l_1}_{e_c}$) using the 
same procedure as described in \S~\ref{ssec:monot}.
\section{Experimental Setup}
\label{sec:expset}
To evaluate our approach, we built a dataset for the cross-language and monolingual pairwise tweet and news article relevance assessment. Also, we used the existing monolingual 
comparisons corpora to compare with other approaches.
\subsection{Corpus Creation}
\label{ssec:cc}
Unavailability of datasets for comparing news articles with the tweets in different languages compelled us to create our own. We created a gold standard dataset for monolingual and 
cross-lingual comparison across collections by acquiring some more tweets and news articles mainly in English and German in the same way as described in \S~\ref{sec:mwc}.

Tweets with a single URL link to any news article are collected and carefully evaluated to see if it does not simply represent the news title or summary. If they only represent news 
title or summary then they are considered to be trivial and are removed. After basic preprocessing, using the keyword ``Grexit'' (the Greece exit of the European Union) around 
18 tweets and 18 news articles (both English and German) are selected for further human evaluation.
\subsection{Human Evaluation}
The goal of the human evaluation is to get pairwise comparison scores between tweets and news. Thus, each participant had to rate a pair of documents with respect to their semantic 
similarity.
Three different annotators who have English(E) and German(G) language skills were chosen for comparing pair of tweets and news based on scores listed in Table~\ref{ss}.
\begin{table}[!ht]
 \small
\begin{center}
\begin{tabular}{lll}
\hline
Score	  &     Type     &   Description  \\ 
\hline
0      &     Dissimilar  &  Tweet and news article are \\
       &		 &  not about same topic. \\
1      &     Related     &  Tweet and news article share topic \\
       &                 &  but important ideas in news is not \\
       &		 &  represented in the tweet. \\
2      &     Similar     &  Tweet and news article are about \\
       &                 &  same topic and important ideas \\
       &                 &  in news is represented in the tweet \\
\hline
\end{tabular}
\end{center}
\caption{\label{ss}Similarity Scores}
\end{table}
At the end, a list of 628 relevance judgments (i.e. 162 between (E)Tweets and (E)News, 162 between (E)Tweets and (G)News and so on) were produced. A significance test with Kendall's $\tau$ 
is computed to test the consistency among user judgments. Results suggested that there is no significant difference in the score pairs of users ($0.05$ 
significance level). Specifically, the results showed that users have an similar understanding of the similarity assessment. To obtain the final score for each pair, similar to SemEval semantic 
similarity tasks\footnote{\texttt{http://ixa2.si.ehu.es/stswiki/index.php/Main\_Page}} arithmetic mean was calculated between all user ratings. We term this resource as 
\textbf{Dataset-1}\footnote{\texttt{http://people.aifb.kit.edu/amo/cicling2017/}}. This dataset provides more fine-grain comparison as compared to other datasets~\cite{krestel:2015} 
that provide only binary relevance. 
\subsection{Other Datasets}
\label{ssec:od}
Evaluation of monolingual comparison is also performed on the other existing resources such as Krestel et al.~\cite{krestel:2015}. This dataset consists of 1600 relevance judgments 
constituting 17 news articles covering different topics with the Tweets labeled as relevant or irrelevant for the each news article. We term this resource as \textbf{Dataset-2}.
\subsection{Evaluation Metrics}
For many pairwise semantic similarity tasks statistical correlation based measures have been used. Here, we use Pearson correlation coefficient ($r$) to evaluate our approaches on 
the dataset we created. While, measures like accuracy is used for other datasets.
\section{Experimental Results}
In this section, we present our experimental results on different datasets with variation in parameters. 
\subsection{Baselines}
Two different baselines are used to compare with our approach.
\subsubsection{Latent Dirichlet Allocation (LDA)}
Most of the earlier research~\cite{zhao:2011,guo:2013} have shown significant interest to compare news and tweets with LDA and its variations. We use the polylingual topic 
model~\cite{mimno:2009} trained on English and German Wikipedia with 100 topics to support multiple languages. Similarity between tweet and news represented as topics vector is 
measured using cosine similarity.
\subsubsection{WTMF-G}
Weighted Textual Matrix Factorization on Graphs (WTMF-G)~\cite{guo:2013} is one of the baseline that compare tweets and news based on a graph connected by hashtags, named entities or temporal 
information. To train the WTMF-G model we used regularization coefficient ($\lambda=20$), weight of missing words as $w_n=0.01$, number of neighbors ($k=4$) and link weights ($\delta=3$) as 
suggested in earlier research. Latent dimension of 100 is used to represent tweet and news, while similarity between them is calculated using cosine similarity.
\subsection{TWE Implementation}
Major parameters that affect training of Glove is the dimensionality of word embeddings and the size of word context window. We choose 25, 50, 100, 200, 400 word embedding dimensions and 
5 words on left and right context window. Similarly, later for learning bilingual word embeddings we used Bilbowa tool\footnote{\texttt{https://github.com/gouwsmeister/bilbowa}} to learn same embedding 
dimensions as former with 5 word left context window and entire English-German Europarl-v7\footnote{\texttt{http://www.statmt.org/europarl/}} as the parallel data. In both cases, count of words less 
than 2 in the entire corpus are discarded.
\subsection{Monolingual Comparison}
\label{ssec:monoc}
Before comparing monolingual news and tweets, we estimate the quality of embedding transformation achieved with \textbf{Monolingual-TWE} by performing similar experiments as 
in \S~\ref{sec:mwc}. The transformation can be either from tweets to news (T2N) or in the opposite orientation (N2T). Though both of them have different transformation, we observed that they 
produce similar t-SNE visualization. Also, there is a slight decrease in distance between common words across collections as compared to without transformation. Average RBO measure using the 
top 5000 frequent terms observed in both tweets and news collections in German and English is recalculated to perceive the refinement. We perceived that there is an improvement of 
24.4\% and 21.2\% for English and German respectively.

\begin{figure*}[!ht]
  \centering{}
    \includegraphics[width=\textwidth]{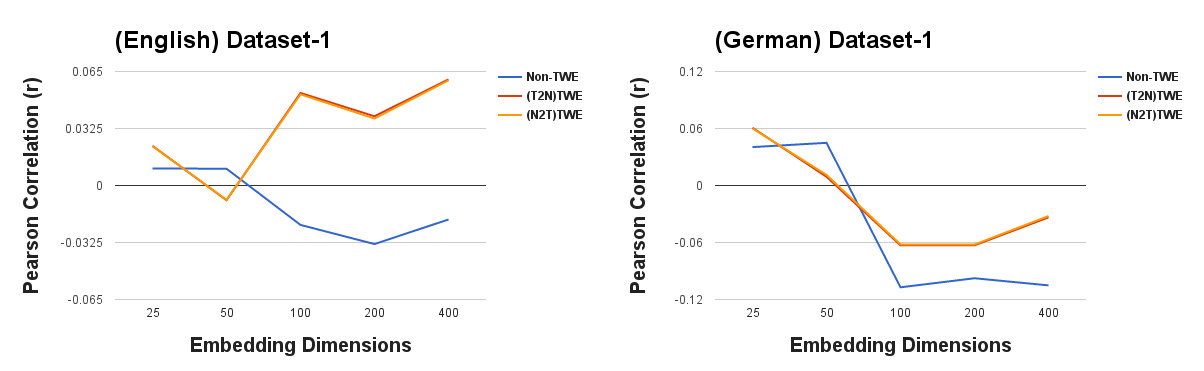}
  \caption{Effect of Embedding Dimensions(Dataset-1)}
  \label{fig:monodata1}
\end{figure*}

\begin{figure*}[!ht]
  \centering{}
    \includegraphics[width=0.5\textwidth]{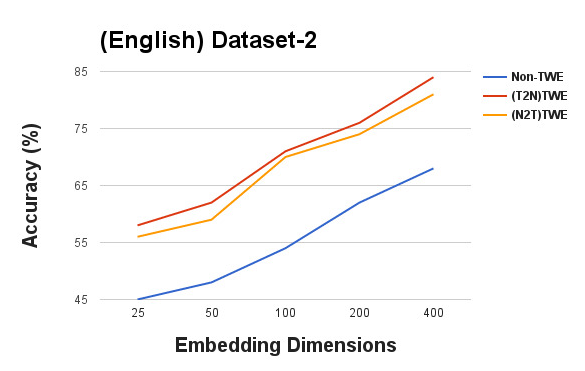}
  \caption{Effect of Embedding Dimensions(Dataset-2)}
  \label{fig:monodata2}
\end{figure*}

Now, tweets and news articles in Dataset-1 and Dataset-2 are represented as the tf-idf weighted average of transformed word embeddings. They are now used as input to 
SVM classifier\footnote{\url{https://www.csie.ntu.edu.tw/~cjlin/libsvm/}} with default parameters to calculate accuracy and to cosine similarity for finding Pearson correlation. 
Furthermore, top performing embedding dimensions are identified based on Pearson correlation and accuracy measures using validation data of the datasets. 
Figure~\ref{fig:monodata1} and Figure~\ref{fig:monodata2} show the comparison of results with ((T2N)TWE and (N2T)TWE) and without (Non-TWE) transformation on different datasets. 
Once the top performing embedding dimensions are identified, testing data is used to compare different approaches with diverse measures in Table~\ref{ourdmono} and Table~\ref{standard}.
\begin{table}[!ht]
  \begin{center}
  \centering
  \small
    \begin{tabular}{clcc}
    \hline
    \multicolumn{2}{c}{Method} & Dim & $r$ \\
    \hline
    {\textbf{German}} & No-Transformation & 400 & -0.1051 \\
		       & LDA-PTM~\cite{mimno:2009} & 100 & 0.0445  \\
		       & WTMF-G~\cite{guo:2013} & 100 & 0.0498\\
		       & (T2N)Monolingual-TWE & 25  & \textbf{0.0607} \\
		       & (N2T)Monolingual-TWE & 25  & 0.0601 \\
    \hline
   {\textbf{English}} & No-Transformation & 400 & -0.1193 \\
		       & LDA-PTM~\cite{mimno:2009} & 100 & 0.0321  \\
		       & WTMF-G~\cite{guo:2013} & 100 &  0.0491 \\
		       & (T2N)Monolingual-TWE & 400  & \textbf{0.0605} \\
		       & (N2T)Monolingual-TWE & 400  & 0.0599 \\
    \hline
  \end{tabular}
  \end{center}
\caption{\label{ourdmono}Monolingual Tweets and News Comparison}
\end{table}
\begin{table}[!ht]
 \small
\begin{center}
\begin{tabular}{lll}
\hline
Method	  &  Dim & Accuracy  \\ 
\hline
LDA-PTM~\cite{mimno:2009}      &  100 &   79.1\% \\
Boosting~\cite{krestel:2015}    &  - &  82.5\%       \\
(T2N)Monolingual-TWE+SVM & 400 & \textbf{83.1\%} \\
(N2T)Monolingual-TWE+SVM & 400 & 81.0\% \\
\hline
\end{tabular}
\end{center}
\caption{\label{standard}Accuracy (English)}
\end{table}
\subsection{Cross-Lingual Comparison}
\label{ssec:clc}
\begin{figure*}[!ht]
  \centering{}
    \includegraphics[width=\textwidth]{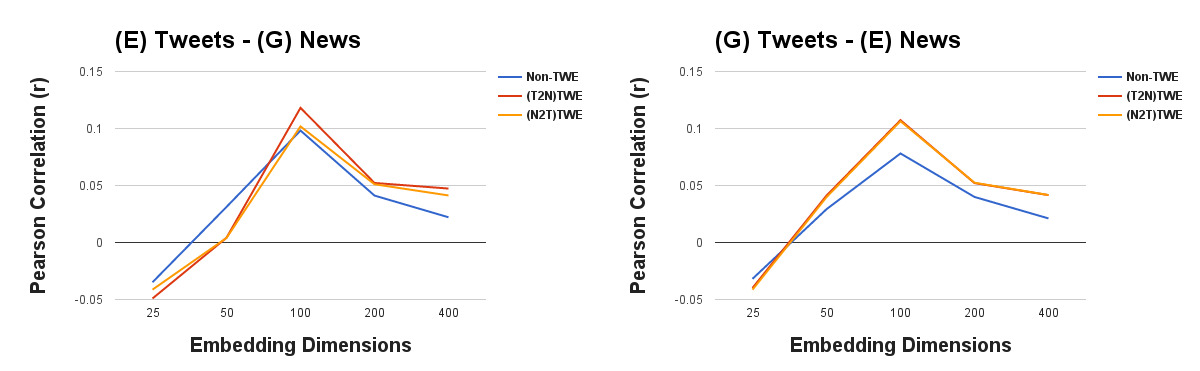}
  \caption{Effect of Embedding Dimensions(Cross-Lingual)}
  \label{fig:cldata1}
\end{figure*}

For the cross-lingual comparison, we follow a similar procedure as in \S~\ref{ssec:monoc}.
Since, news word embeddings incorporate bilingual information from both German and English, calculation of RBO measure between tweets and news without transformation is not appropriate. 
Hence, we calculate RBO measure after transformation to verify that it satisfies minimum threshold of $0.328$, which in general fetch satisfactory results~\cite{tan:2015}. Now to compare 
tweets and news belonging to the dataset listed in \S~\ref{ssec:cc} across languages, we estimate the top performing embedding dimension based on Pearson correlation measure using 
the validation data of the dataset. Figure~\ref{fig:cldata1} show the comparison of results with (TWE) and without (Non-TWE) transformation. Once the top performing embedding dimension is 
identified, testing data is used to compare different approaches as provided in Table~\ref{ourdcl}.
\begin{table}[!ht]
  \begin{center}
  \small
    \centering
    \begin{tabular}{clc}
    \hline
    \multicolumn{2}{c}{Method} & $r$ \\
    \hline
    {\textbf{(E)Tweets - (G)News}} & LDA-PTM~\cite{mimno:2009} & 0.0821 \\
		       & (T2N)Cross-Lingual-TWE & \textbf{0.1181} \\
		       & (N2T)Cross-Lingual-TWE & 0.1018 \\
    \hline
    {\textbf{(G)Tweets - (E)News}}  & LDA-PTM~\cite{mimno:2009} & 0.0765  \\
		       & (T2N)Cross-Lingual-TWE & \textbf{0.1073} \\
		       & (N2T)Cross-Lingual-TWE & 0.1064 \\
    \hline
   \end{tabular}
   \end{center}
\caption{\label{ourdcl}Cross-Lingual Tweets and News Comparison With 100-Dimensions}
\end{table}
\section{Discussion}
We start our analysis with results observed in the Table~\ref{ourdmono}. It can be comprehended that the Monolingual-TWE (either T2N or N2T) achieved an commendable improvement 
over other approaches. However, the values for Pearson correlation are low and can be associated to the fact that Tweets and news are inherently very different and achieving high level of 
pairwise similarity is a complex task. But for accuracy assessment, which is mostly seen from the perspective of a classification task there is clear improvement 
over other approaches by using transformed embeddings as features. Table~\ref{standard} shows that T2N achieved better performance as compared to N2T. 

Although aforementioned analysis is perceived on a small dataset. The results show a promising direction to use Monolingual-TWE which can easily scale with the size of common 
vocabulary across collections. Thus giving a possibility to improve or sustain the accuracy and Pearson correlation values on larger datasets. 

Similar observations can be enunciated about cross-lingual-TWE. Given the complexity associated with finding pairwise relevance between tweets and cross-language news, we compared only LDA 
based approaches with cross-lingual-TWE. It can be comprehended from Table~\ref{ourdcl} that T2N outperformed LDA-PTM with notable improvement. Although it may not be significant, these 
results only show preliminary examination to perceive research in this direction.
\section{Conclusion and Future Work}
\label{sec:conc}
In this paper, we focused on mapping tweets with monolingual and cross-lingual news by transforming their word embeddings closer to each other, thus bridging the lexical and word 
usage gap across collections. In future, we aim to improve the quality of results with more sophisticated approaches.
\bibliography{llncs}

\begin{thebibliography}{10}

\bibitem{kwak:2010}
Kwak, H., Lee, C., Park, H., Moon., S.:
\newblock What is twitter, a social network or a news media?.
\newblock In: Proceedings of WWW., ACM (2010)  591--600

\bibitem{zhao:2011}
Zhao, W.X., Jiang, J., Weng, J., He, J., Lim, E.P., Yan, H., Li., X.:
\newblock Comparing twitter and traditional media using topic models.
\newblock In: Advances in Information Retrieval., Springer Berlin Heidelberg
  (2011)  338--349

\bibitem{guo:2013}
Guo, W., Li, H., Ji, H., Diab., M.T.:
\newblock Linking tweets to news: A framework to enrich short text data in
  social media.
\newblock In: Proceddings of ACL. (2013)  239--249

\bibitem{krestel:2015}
Krestel, R., Werkmeister, T., Wiradarma, T.P., Kasneci., G.:
\newblock Tweet-recommender: Finding relevant tweets for news articles.
\newblock In: Proceedings of WWW, ACM (2015)  53--54

\bibitem{pennington:2014}
Pennington, J., Socher, R., Manning., C.D.:
\newblock Glove: Global vectors for word representation.
\newblock In: Proceedings of EMNLP. (2014)  1532--1543

\bibitem{gouws:2014}
Gouws, S., Bengio, Y., Corrado., G.:
\newblock Bilbowa: Fast bilingual distributed representations without word
  alignments.
\newblock In: arXiv preprint arXiv:1410.2455. (2014)

\bibitem{wang:2008}
Wang, C., Mahadevan., S.:
\newblock Manifold alignment using procrustes analysis.
\newblock In: Proceedings of ICML, ACM (2008)  1120--1127

\bibitem{tan:2015}
Tan, L., Zhang, H., Clarke, C.L., Smucker., M.D.:
\newblock Lexical comparison between wikipedia and twitter corpora by using
  word embeddings.
\newblock In: Proceedings of ACL. (2015)

\bibitem{ritter:2012}
Ritter, A., Etzioni, O., Clark., S.:
\newblock Open domain event extraction from twitter.
\newblock In: Proceedings of KDD. (2012)  1104--1112

\bibitem{thelwall:2011}
Thelwall, M., Buckley, K., Paltoglou., G.:
\newblock Sentiment in twitter events.
\newblock Journal of the American Society for Information Science and
  Technology. \textbf{62} (2011)  406--418

\bibitem{paul:2011}
Paul, M.J., Dredze., M.:
\newblock You are what you tweet: Analyzing twitter for public health.
\newblock In: Proceedings of ICWSM. (2011)  265--272

\bibitem{ratkiewicz:2011}
Ratkiewicz, J., Conover, M., Meiss, M., Goncalves, B., Flammini, A., Menczer.,
  F.:
\newblock Detecting and tracking political abuse in social media.
\newblock In: Proceedings of ICWSM. (2011)

\bibitem{crooks:2013}
Crooks, A., Croitoru, A., Stefanidis, A., Radzikowski., J.:
\newblock \#earthquake: Twitter as a distributed sensor system.
\newblock Transactions in GIS. \textbf{17(1)} (2013)  124--147

\bibitem{abel:2011}
Abel, F., Gao, Q., Houben, G.J., Tao., K.:
\newblock Analyzing user modeling on twitter for personalized news
  recommendations.
\newblock In: Proceddings of UMAP. (2011)  1--12

\bibitem{jin:2011}
Ou, J., Liu, N.N., Zhao, K., Yu, Y., Yang., Q.:
\newblock Transferring topical knowledge from auxiliary long texts for short
  text clustering.
\newblock In: Proceddings of CIKM., ACM (2011)  775--784

\bibitem{coll:2011}
Collobert, R., Weston, J., Bottou, L., Karlen, M., Kavukcuoglu, K., Kuksa., P.:
\newblock Natural language processing (almost) from scratch.
\newblock The Journal of Machine Learning Research \textbf{12} (2011)
  2493--2537

\bibitem{alrfou:2013}
Al-Rfou, R., Bryan, P., Steven., S.:
\newblock Polyglot: Distributed word representations for multilingual nlp.
\newblock In: Proceedings of CoNLL, ACL (2013)  183--192

\bibitem{ramon:2015}
Ramon, A.F., Amir, S., Lin, W., Silva, M., Trancoso., I.:
\newblock Learning word representations from scarce and noisy data with
  embedding sub-spaces.
\newblock In: Proceedings of ACL. (2015)

\bibitem{kim:2015}
Kim, J., Rousseau, F., Vazirgiannis., M.:
\newblock Convolutional sentence kernel from word embeddings for short text
  categorization.
\newblock In: Proceedings of EMNLP. (2015)

\bibitem{van:2008}
der Maaten, L.V., Hinton., G.:
\newblock Visualizing data using t-sne.
\newblock The Journal of Machine Learning Research \textbf{9} (2008)
  2579--2605

\bibitem{webber:2010}
Webber, W., Moffat, A., Zobel., J.:
\newblock A similarity measure for indefinite rankings.
\newblock ACM Transactions on Information Systems (TOIS). \textbf{4} (2010)

\bibitem{schonemann:1966}
Sch{\"o}nemann, P.H.:
\newblock A generalized solution of the orthogonal procrustes problem.
\newblock Psychometrika. \textbf{31(1)} (1966)  1--10

\bibitem{mimno:2009}
Mimno, D., Wallach, H.M., Naradowsky, J., Smith, D.A., McCallum., A.:
\newblock Polylingual topic models.
\newblock In: Proceedings of EMNLP, ACL (2009)  880--889

\end{thebibliography}
\bibliographystyle{splncs}
\end{document}